\documentclass{ecai} 

\usepackage{latexsym}
\usepackage{amssymb}
\usepackage{amsmath}
\usepackage{amsthm}
\usepackage{booktabs}
\usepackage{enumitem}
\usepackage{graphicx}
\usepackage{color}
\usepackage{multirow}
\usepackage{subfigure}

\newcommand{\BibTeX}{B\kern-.05em{\sc i\kern-.025em b}\kern-.08em\TeX}

\begin{document}


\begin{frontmatter}


\paperid{511} 


\title{TIGER: Temporally Improved Graph Entity Linker}


\author[A]{\fnms{Pengyu}~\snm{Zhang}\orcid{0000-0001-5111-4487}\thanks{Corresponding Author. Email: p.zhang@uva.nl.}}
\author[A]{\fnms{Congfeng}~\snm{Cao}\orcid{0000-0001-9011-3807}}
\author[A]{\fnms{Paul}~\snm{Groth}\orcid{0000-0003-0183-6910}} 

\address[A]{University of Amsterdam, Amsterdam, The Netherlands}


\begin{abstract}
Knowledge graphs change over time, for example, when new entities are introduced or entity descriptions change. This impacts the performance of entity linking, a key task in many uses of knowledge graphs such as web search and recommendation. Specifically, entity linking models exhibit temporal degradation - their performance decreases the further a knowledge graph moves from its original state on which an entity linking model was trained. To tackle this challenge, we introduce \textbf{TIGER}: a \textbf{T}emporally \textbf{I}mproved \textbf{G}raph \textbf{E}ntity Linke\textbf{r}. By incorporating structural information between entities into the model, we enhance the learned representation, making entities more distinguishable over time. The core idea is to integrate graph-based information into text-based information, from which both distinct and shared embeddings are based on an entity's feature and structural relationships and their interaction. Experiments on three datasets show that our model can effectively prevent temporal degradation, demonstrating a 16.24\% performance boost over the state-of-the-art in a temporal setting when the time gap is one year and an improvement to 20.93\% as the gap expands to three years. The code and data are made available at \url{https://github.com/pengyu-zhang/TIGER-Temporally-Improved-Graph-Entity-Linker}.
\end{abstract}

\end{frontmatter}


\section{Introduction}

A Knowledge Graph (KG) is a structured representation of facts, consisting of entities, relationships between them and their attributes \cite{9416312}. KGs such as DBpedia \cite{10.1007/978-3-540-76298-0_52}, YAGO \cite{10.1145/1242572.1242667}, and Wikidata \cite{10.1145/2629489}, not only capture the relationships between entities but also contain rich textual attributes. These and other KGs play an important role in web applications such as recommendation systems \cite{10.1145/3477495.3532025} and question answering \cite{9770789}. However, performance on the above applications is frequently limited by the ambiguity of entities mentioned in the text. For example, `apple' could refer to a fruit or a multinational technology company.

\begin{figure}[t]
\centering
\subfigure[Entity linking with just textual information.]{\includegraphics[width=0.85\linewidth]{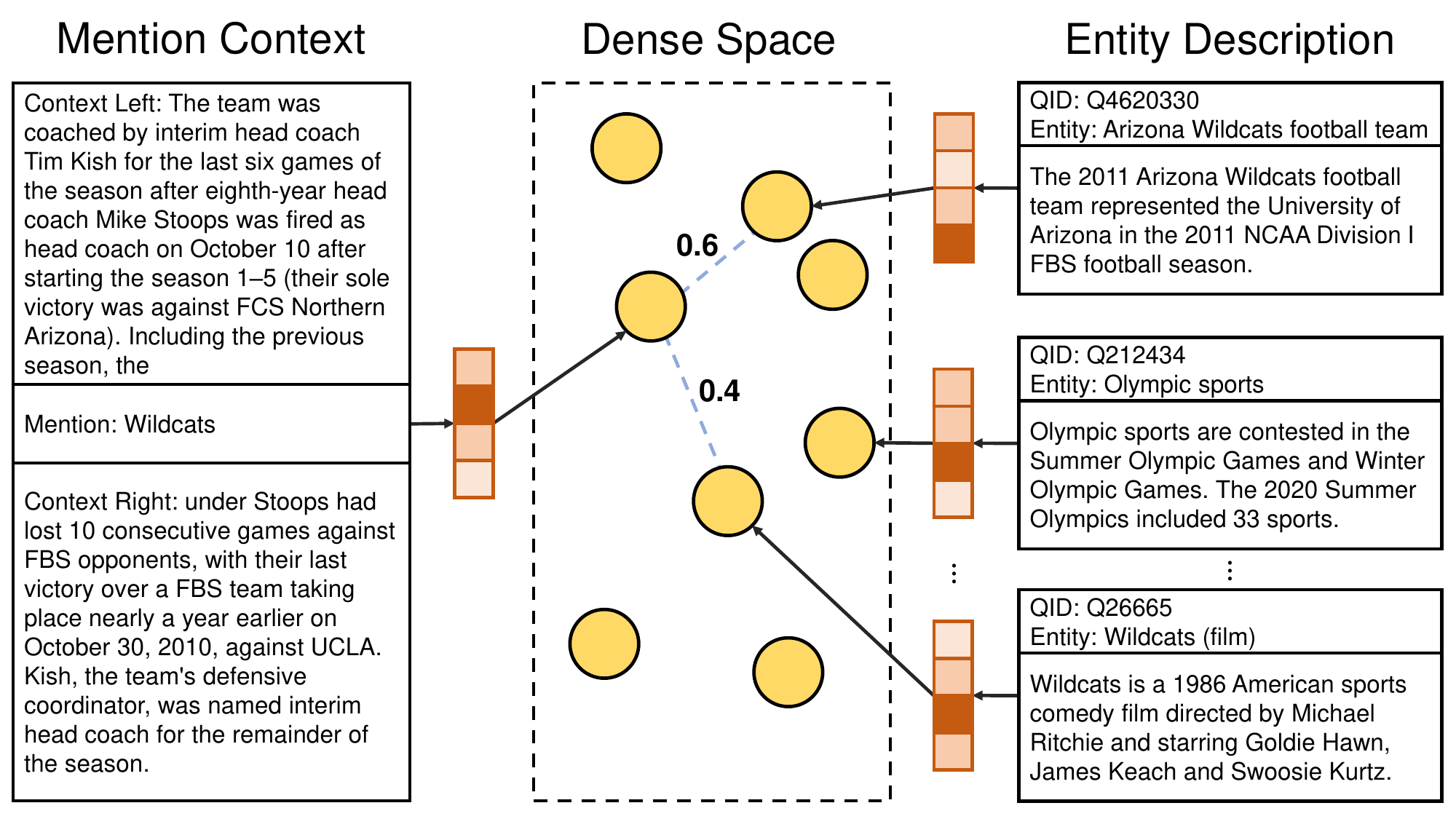}}
\subfigure[Entity linking with both textual and structural data, where edge colors denote different entity relationships.]{\includegraphics[width=0.85\linewidth]{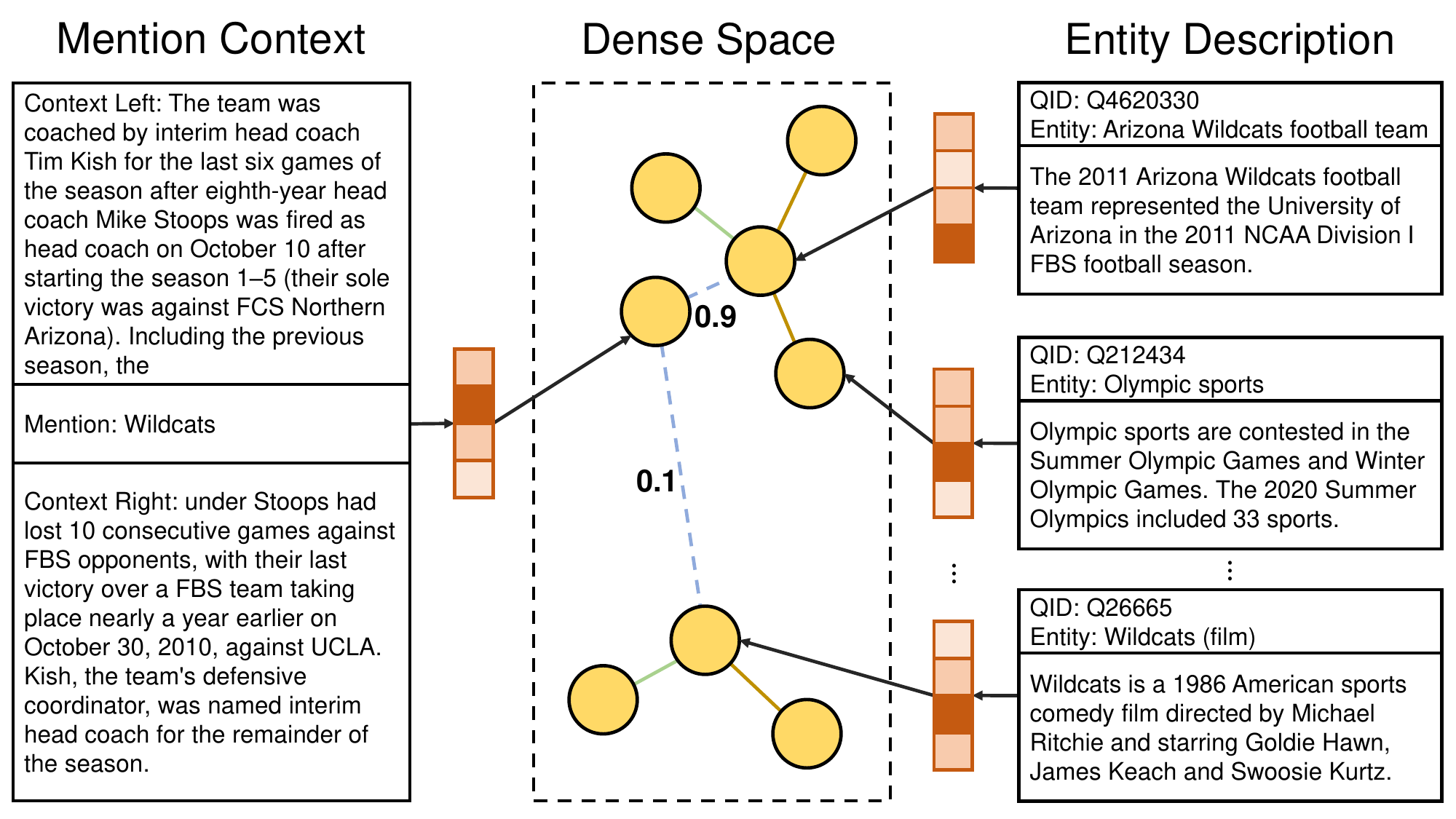}}
\caption{In this example, the mention `Wildcats' could refer to Arizona Wildcats football team, or Wildcats (film) entities. We can learn better entity representations by including information from the graph structure.} \label{figure1}
\end{figure}

Hence, the task of Entity Linking (EL) has emerged as a vital step in both producing and using KGs. EL aims to connect mentions of entities in text to their corresponding entities in a KG. Even though there are many works in EL, only a few consider the time aspect. In real-world scenarios, KGs evolve over time \cite{tietz2020challenges}, existing entities (continual entities) may change their meanings over time due to societal development, and previously non-existent entities (new entities) may appear. For example, entity descriptions from Wikipedia constantly evolve (e.g., the most frequent meaning of the term `corona' changed around 2020). New entities emerge (e.g., `COVID-19' was added to Wikidata in 2020). Neglecting the evolution of entities can lead to less accurate EL.
 
To tackle this challenge, \cite{zaporojets2022tempel} proposed a unique dataset, TempEL, to explore the temporal evolution aspect of the EL task. The dataset consists of 10 yearly snapshots, evenly distributed, from English Wikidata entities, spanning from January 1, 2013, to January 1, 2022. However, despite the model presented with the TempEL dataset achieving impressive results across snapshots, it still suffers from temporal degradation. When trained on data from time $t_1$ and tested on data from time $t_2$, performance declines as the gap between the two timestamps widens. Such degradation may constrain the effectiveness of EL models in dynamically changing real-world contexts.

In response to the challenge of temporal degradation, we hypothesize that the relationships between entities can serve as vital information in the EL task, as shown in Figure~\ref{figure1}. The simultaneous attention to both textual and structural information could substantially enhance the accuracy and robustness of EL in the face of evolving temporal contexts. Because entities are similar to nodes in the network, their meanings are influenced by their intrinsic properties and the nature of their connections to other nodes. Consider an entity ambiguously labeled `apple' in two distinct periods. In the earlier period, surrounding nodes and edges might be related to `orchards' and `fruit'. At a later time, connections might be made to `technology' and `innovation'. Through the graph's structural context, we can verify that the former is likely referencing the fruit, while the latter implies the tech company. Hence, we contend that we can learn better entity representations by including information from the graph structure, resolving unwanted ambiguities. 

To incorporate structural information about entities, we proposed \textbf{TIGER}, a \textbf{T}emporally \textbf{I}mproved \textbf{G}raph \textbf{E}ntity Linke\textbf{r}. By including the structural information in our model, entities are better described. As a result, each entity's representation becomes more distinct and easier to differentiate from other entities. Our contributions are summarized as follows:

\begin{itemize}
\item The Graph-TempEL dataset that includes relationships from the Wikidata5M \cite{wang2021kepler} dataset to study the time-evolving aspect of entity linking tasks.
\item A novel entity linking model that adaptively combines both text and graph information.
\item Extensive experiments on three entity linking datasets show notable improvement of our model over related approaches.
\end{itemize}

\section{Related Work}

\paragraph{Entity Linking.} Entity Linking (EL) sometimes called Wikification, is the connecting of mentions of entities in the text to a knowledge base and is a widely studied topic in NLP. We refer the reader to \cite{sevgili2022neural} for a detailed survey of the topic. Here, we focus on key challenges faced by current EL models. One of these challenges is linking textual mentions to unseen entities, known as zero-shot learning \cite{10.5555/1620163.1620172}. For instance, \cite{wu2019zero} introduces a conceptually two-stage, highly effective BERT-based zero-shot EL model called BLINK. \cite{bhargav-etal-2022-zero} proposed a neuro-symbolic, multi-task learning approach to mitigate the problem of diminishing returns. They improved BLINK's performance with much less data by exploiting auxiliary information about entity types. To address the issue of an entity not being present in the knowledge base, NASTyLinker \cite{10.1007/978-3-031-33455-9_11} clusters mentions and entities using dense representations from Transformers and, if multiple entities are assigned to a single cluster, it resolves conflicts by calculating transitive mention-entity affinities. Building on the idea of understanding intricate relationships between entities, \cite{van-erp-groth-2020-towards} introduced a novel concept of representing entities in multi-dimensional spaces, which could further refine the EL process. Furthermore, \cite{sui-etal-2022-improving} proposes a hierarchical multi-task model to extract ultra-fine type information that can help to learn contextual commonality and improve their generalization ability to tackle the overfitting problem. An additional challenge related to unseen entities is the problem addressed in this paper, temporal EL, where both unseen and changing entities must be linked \cite{zaporojets2022tempel}.

Several existing studies have sought to combine graph vectors with textual content to address the zero-shot problem. Among them, KG-ZESHEL \cite{10.1145/3460210.3493549} stands out for its innovative approach. Their approach lies in integrating graph vectors, which provide a route to combine textual and graph knowledge from knowledge graphs. This information fusion could enhance the model's ability to resolve ambiguities and improve EL accuracy. However, the study primarily focuses on the zero-shot scenario in EL, overlooking the challenges of temporal degradation. Furthermore, KG-ZESHEL does not fully exploit unique and shared features across different graphs or relationships.

\paragraph{GNN-based Knowledge Graph Models.} Graph Neural Networks (GNNs) integrate the topological and attribute information inherent in graph data through deep neural networks, thereby generating more refined node feature representations \cite{9831453}. Recently, several studies have focused on using node features derived from graph representation learning in the context of knowledge graphs. For instance, the Contextualized Graph Attention Network (CGAT) \cite{9439972} effectively leverages both local and non-local graph context information of KG entities. Essential entities for a target entity are extracted from the entire KG via a biased random walk, thereby incorporating non-local context within the KG. DSKReG \cite{10.1145/3459637.3482092} proposed learning the relevance distribution of associated items from knowledge graphs and sampling relevant items by this distribution to prevent the exponential growth of a node's receptive field. The work in \cite{10.1145/3366423.3380197} creates a dense, high-coverage semantic subgraph by linking question entity nodes to candidate entity nodes via text sentences from Wikipedia.

\paragraph{Temporal Degradation.} Temporal change on the web has been well documented both for structured \cite{abian2022analysis} and unstructured information \cite{10.1145/3366423.3380113}. However, temporal dependency in models is often overlooked. The common assumption is that once a model reaches the desired level of quality, it can be deployed without requiring further updates or retraining \cite{vela2022temporal}. This assumption, however, may not hold true for tasks involving KGs, where entities evolve over time. The impact of temporal variation of KGs on model performance has been shown in several use cases ranging from online shopping \cite{10.1145/3485447.3512064} and internet of things \cite{10.1145/3447032}. The TempEL paper highlights the same need to address temporal degradation for EL, which we do here.

\section{Task Formulation and Definition}

\textbf{Entity Linking (EL).} The EL task takes a given text document $\mathbf{D}$ as input, which comprised of a list of tokens $\left[w_1, \ldots, w_r\right]$, where $r$ indicates the document's length. Within this document, there exists a list of entity mentions $\mathbf{M}_{\mathbf{D}}$ containing $n$ distinct elements $\left[m_1, \ldots, m_n\right]$, where each mention $m_{i}$ corresponds to a span of continuous tokens in $\mathbf{D}$, represented as $m_i=\mathbf{D}\left[x, y\right]$. The model subsequently yields a list of mention-entity pairs $\left\{\left(m_i, e_i\right)\right\}_{i \in[1, n]}$. Every entity $e_i$ correlates with an entry within a comprehensive knowledge base (KB), such as Wikipedia. It is assumed that both the title and description of these entities are available, a standard premise in EL \cite{logeswaran-etal-2019-zero}.

\textbf{Graph.} A graph is defined as $G=(V, E)$, where $V$ is the set of $N$ nodes (i.e., entities) $\left\{v_1, v_2, \cdots, v_N\right\}$. $E$ is the set of $M$ edges (i.e., relations) represented as $\left\{e_1, e_2, \cdots, e_M\right\}$,  where each $e_i$ is a pair of nodes from $V$, such as $e_{i}=(v_{a}, v_{b})$. A graph is termed homogeneous when all its nodes and edges belong to the same type, where the number of node types is 1, and the number of edge types is also 1 \cite{9780569}.

\section{Dataset Construction}

We combined the TempEL\footnote{https://cloud.ilabt.imec.be/index.php/s/RinXy8NgqdW58RW} dataset, a benchmark for temporal EL, with Wikidata5M\footnote{https://deepgraphlearning.github.io/project/wikidata5m} to explore the benefits of structured knowledge graphs. Our resultant dataset has five segments: two text-based (\textbf{entity description} and \textbf{mention context}) and three graph-based (\textbf{structure graph}, \textbf{feature graph}, and \textbf{feature matrix}). The construction process is shown in Figure~\ref{figure2}. We make the dataset available in the supplementary material \cite{zhang_2024_12790573}. We now walk through each step. 

\begin{figure}[th]
\centering
\includegraphics[width=0.85\linewidth]{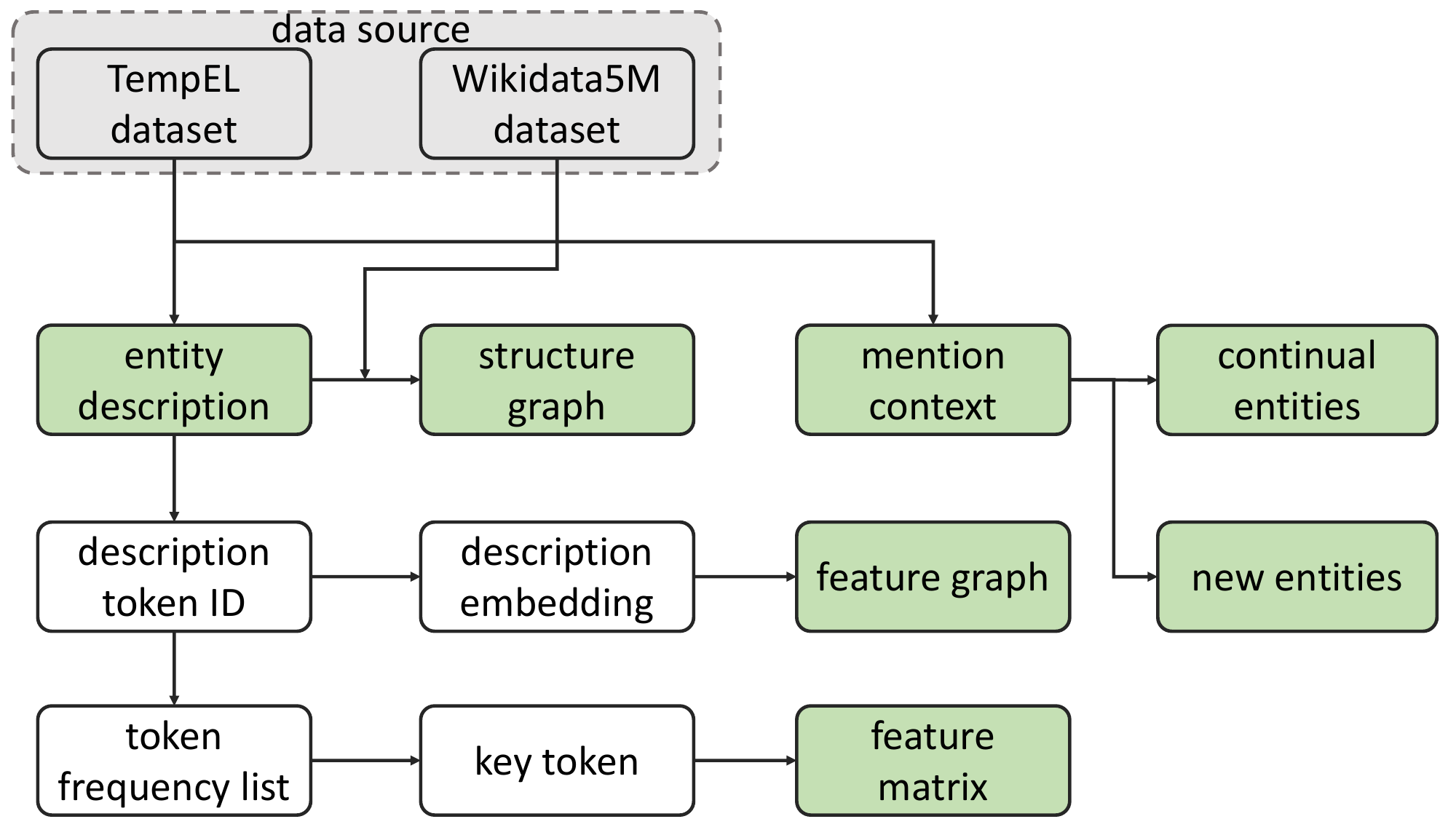}
\caption{The dataset construction process. We use Wikidata5M to extend TempEL with \textit{strutured graph} representations. The green section represents the input to our model.}\label{figure2}
\end{figure}

First, we categorized each year of data from the TempEL dataset into entity descriptions and mention context parts based on the year. The entity description comprises the title, text, document ID, and, importantly, the unique ID of the entity (its QID). The mention context consists of context left, context right, mention, label, QID, and category.

Second, we create a structure graph based on the relationship in the Wikidata5M dataset and the entity IDs in the TempEL dataset. There are numerous relationships among entities in the Wikidata5M dataset. To filter down the number of relationships, we matched these relationships' entity IDs (also QIDs) with the QID in the entity descriptions from the TempEL dataset. We will keep the relationship if both QIDs are in a relationship in the Wikidata5M data and are present in the existing entity description. The structure graph is an \( n \times n \) adjacency matrix, where \( n \) represents the total number of entities in the dataset. Each row indicates whether an entity has a connection with another entity. The adjacency matrix is made up of 0s and 1s. If entity \( i \) and entity \( j \) are connected, the value in the \( i^{th} \) row and \( j^{th} \) column of the matrix is 1; otherwise, it is 0.

Third, we built the feature graph using the embeddings from entity descriptions. We employed the pre-trained bert-base-uncased model to embed the entity description's textual information associated with the `text' key. By accessing the embedded information for each entity in the dataset, we established a $k$NN graph based on these entities, which we refer to as the feature graph. This graph highlights the connections between entities based on their entity descriptions. The feature graph is also an \( n \times n \) adjacency matrix, where \( n \) represents the total number of entities in the dataset. Each row indicates whether an entity has a connection with other entities. If entity \( i \) and entity \( j \) are connected, the value in the \( i^{th} \) row and \( j^{th} \) column of the matrix is 1; otherwise, it is 0.

Fourth, we constructed a feature matrix representing each entity based on the tokens from entity descriptions in the dataset. After getting the token IDs for each entity using the pre-trained bert-base-uncased model, we filtered all token IDs based on their frequency of occurrence. We retained those token IDs that appeared between 46 and 200 times. We discarded highly frequent token IDs since these tokens, such as `is,' `an,' `the,' and other common words, do not offer meaningful differentiation among entities. Also, the less frequent token IDs were removed due to the possibility of them being meaningless noise or random codes, and including an excess of these rare tokens would make the matrix too sparse, slowing down computation. The final feature matrix is an \( n \times m \) dimensional matrix composed of 0s and 1s. Here, \( n \) represents the total number of entities in the dataset, while \( m \) is the number of retained token IDs. If the data in the \( i^{th} \) row and \( j^{th} \) column of the matrix is 1, it indicates that entity \( i \) contains the \( j^{th} \) token.

Finally, we generate distinct mention context subsets from all available mention context samples. Using the `category' in each sample as the standard, we further divided the training set into two sub-training sets: `Continual entities (existing entities in previous years)' and `New entities (newly appeared, previously non-existent entities).'

\section{Approach}

\begin{figure*}[th]
\centering
\includegraphics[width=\linewidth]{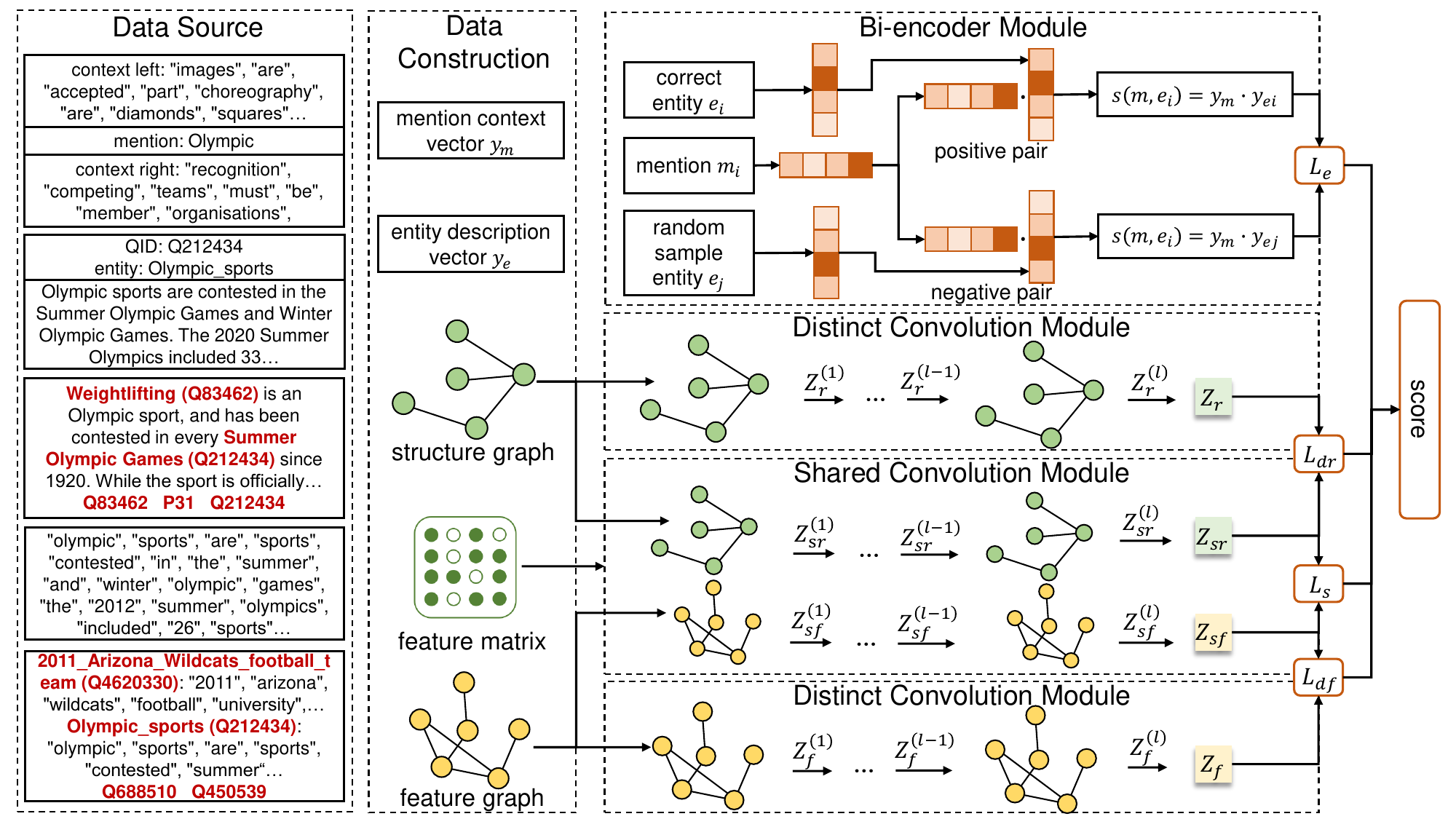}
\caption{The proposed TIGER model adaptively integrates text data (mention context and entity descriptions) with graph data (structural graphs, feature matrices, and feature graphs) to enhance temporal accuracy. The model employs a Shared Convolution Module to learn common features and two Distinct Convolution Modules to capture unique features. Additionally, loss functions are used to emphasize these distinctions.}
\label{figure3}
\end{figure*}

Figure~\ref{figure3} illustrates our model's framework. The core concept is combining text-based information (entity description, mention, and its context at \( t_1 \)) with graph-based data (structure graph, feature graph, and feature matrix at \( t_1 \)) during training.  This integration not only enhances accuracy at \( t_1 \) but also at subsequent times like \( t_2 \). For inference, the model solely relies on text-based information, including entity description, mention, and context.

We now walk through the framework. First, the bi-encoder module employs two separate BERT transformers to transform mention context and entity description into dense vectors \( y_m \) and \( y_e \). Entity candidates are scored via the dot product of these vectors. We introduce \( L_e \) to maximize the correct entity's score against randomly sampled entities.

Second, we input the pre-constructed structure graph, feature graph, and feature matrix into the Distinct and Shared Convolution Modules. Understanding the shared and unique features in both graphs, we use a shared-parameter strategy to derive common embeddings labeled as \( \mathbf{Z}_{sr} \) and \( \mathbf{Z}_{sf} \). A consistency loss \( L_s \) is introduced to emphasize shared features. Meanwhile, distinction losses \( L_{dr} \) and \( L_{df} \) are used to retain the distinctiveness of \( \mathbf{Z}_r \) from \( \mathbf{Z}_{sr} \) and \( \mathbf{Z}_f \) from \( \mathbf{Z}_{sf} \), respectively. Lastly, all loss functions are unified for joint optimization.

\subsection{Bi-encoder Module}

\textbf{Mention Representation.} Following \cite{wu2019zero}, the mention representation $\tau_m$ is constructed from word-pieces of the surrounding context and the mention:
\begin{equation}
[\mathrm{CLS}] \operatorname{ctxt}_l\left[\mathrm{M}_s\right] \text { mention }\left[\mathrm{M}_e\right] \operatorname{ctxt}_r[\mathrm{SEP}]
\end{equation}
where ctxt$_l$, ctxt$_r$ denote word-pieces tokens before and after the mention, and $\left[\mathrm{M}_s\right]$, $\left[\mathrm{M}_e\right]$ tag the mention. The input's maximum length is set to 128, consistent with the BLINK model.

\textbf{Entity Representation.} The representation $\tau_e$ consists of word-pieces of the entity title and its description:
\begin{equation}
[\mathrm{CLS}] \text { title }\left[\mathrm{ENT}\right] \text { description }\left[\mathrm{SEP}\right]
\end{equation}
where $\left[\mathrm{ENT}\right]$ separates the title and description.

\textbf{Encoding.} Using the bi-encoder architecture from \cite{wu2019zero}, we encode descriptions into vectors $y_e$ and $y_m$:
\begin{align}
\boldsymbol{y}_{\boldsymbol{m}} & =\operatorname{red}\left(T_1\left(\tau_m\right)\right) \\
\boldsymbol{y}_{\boldsymbol{e}} & =\operatorname{red}\left(T_2\left(\tau_e\right)\right)
\end{align}
Here, $T_1$ and $T_2$ are transformers, and red(.) reduces the sequence of vectors into a single vector.

\textbf{Scoring.} Entity candidate scores are computed via dot-product:
\begin{equation}
s\left(m, e_i\right)=\boldsymbol{y}_{\boldsymbol{m}} \cdot \boldsymbol{y}_{\boldsymbol{e}_{\boldsymbol{i}}}
\end{equation}

\subsection{Relation Convolution Module}

We input the structure graph (from entity relationships), feature graph (from entity features), and feature matrix (based on token frequencies in entity descriptions) into the Distinct and Shared Convolution Modules.

\textbf{Distinct Convolution Module.} We believe that our model can extract valuable insights from different entity relationships. By feeding the adjacency matrices \(\mathbf{A}_f\) and \(\mathbf{A}_r\) based on entity structure and feature graphs into Distinct Convolution Modules, we obtain two specific embeddings \(\mathbf{Z}_f\) and \(\mathbf{Z}_r\).

Utilizing the pre-constructed feature matrix \(\mathbf{X}\) and adjacency matrix \(\mathbf{A}_f\) that based on feature graph, the output of the \(l\)-th layer, \(\mathbf{Z}_f^{(l)}\), can be represented as:
\begin{equation}
\mathbf{Z}_f^{(l)} = \operatorname{ReLU}\left(\tilde{\mathbf{D}}_f^{-\frac{1}{2}} \tilde{\mathbf{A}}_f \tilde{\mathbf{D}}_f^{-\frac{1}{2}} \mathbf{Z}_f^{(l-1)} \mathbf{W}_f^{(l)}\right)
\end{equation}
with ${\mathbf{W}}_f^{(l)}$ as the weight matrix for the $l$-th GCN layer, initial \(\mathbf{Z}_f^{(0)} = \mathbf{X}\). $\tilde{\mathbf{A}}_f=\mathbf{A}_f+\mathbf{I}_f$ and $\tilde{\mathbf{D}}_f$ is the diagonal degree matrix of $\tilde{\mathbf{A}}_f$. The final layer output is denoted as \(\mathbf{Z}_f\). In this way, we can learn the entities embedding which captures the specific information $\mathbf{Z}_f$ in feature space.

Similarly, using the adjacency matrix \(\mathbf{A}_r\) that based on structure graph and feature matrix \(\mathbf{X}\), the output embedding \(\mathbf{Z}_r\) can be calculated in the same way as in feature graph.

\textbf{Shared Convolution Module.} The feature graph and structure graph are not entirely independent. In the Entity Linking (EL) task, entity feature may be correlated with the feature graph or in structure graph or both of them, which is difficult to know beforehand. Therefore, we not only need to extract the embedding in these two graph, but also to extract the shared information. To address this, we use Shared Convolution Module
with parameter sharing strategy.

Using GCN on the feature adjacency matrix \(\mathbf{A}_f\), the embedding \(\mathbf{Z}_{sf}^{(l)}\) is:
\begin{equation}
\mathbf{Z}_{s f}^{(l)} = \operatorname{ReLU}\left(\tilde{\mathbf{D}}_f^{-\frac{1}{2}} \tilde{\mathbf{A}}_f \tilde{\mathbf{D}}_f^{-\frac{1}{2}} \mathbf{Z}_{s f}^{(l-1)} \mathbf{W}_s^{(l)}\right)
\end{equation}
with \(\mathbf{W}_s^{(l)}\) as the $l$-th GCN layer weight matrix and initial \(\mathbf{Z}_{sf}^{(0)} = \mathbf{X}\).

Similarly, using the adjacency matrix \(\mathbf{A}_r\) that based on structure graph and feature matrix \(\mathbf{X}\), the output embedding \(\mathbf{Z}_{sr}\) can be calculated in the same way as in feature graph.

\subsection{Objective Function}

In order to achieve high EL accuracy, we use the EL loss function $L_e$, distinct convolution loss function $L_{dr}$ and $L_{df}$, shared convolution loss function $L_s$.

\textbf{EL Loss Function $L_e$.} The objective is to train the network such that it maximizes the score of the correct entity compared to the other entities from the same batch. Specifically, for each training pair $\left(m_i, e_i\right)$ within a batch of $N$ pairs, the loss is given by:
\begin{equation}
{L}_e\left(m_i, e_i\right)=-s\left(m_i, e_i\right)+\log \sum_{j=1}^N \exp \left(s\left(m_i, e_j\right)\right)
\end{equation}

\textbf{Shared Convolution Loss Function $L_s$.} Given the output embeddings $\mathbf{Z}_{sr}$ and $\mathbf{Z}_{sf}$ from the GCN with shared weight matrices, the aim is to capture the similarity across $n$ entities. The shared convolution loss ensures that the similarity matrices for both embeddings are consistent, resulting in the following constraint:
\begin{equation}
{L}_s=\left\|\left(\mathbf{Z}_{sr} \cdot \mathbf{Z}_{sr}^T\right) - \left(\mathbf{Z}_{sf} \cdot \mathbf{Z}_{sf}^T\right)\right\|_F^2
\end{equation}

\textbf{Distinct Convolution Loss Functions $L_{dr}$ and $L_{df}$.} To ensure the embeddings $\mathbf{Z}_r$ and $\mathbf{Z}_{sr}$, derived from the same adjacency matrix $\mathbf{A}_r$, capture distinct information, we employ the Hilbert-Schmidt Independence Criterion (HSIC) \cite{10.1145/1273496.1273600}. The HSIC measure is defined as:
\begin{equation}
HSIC\left(\mathbf{Z}_{r}, \mathbf{Z}_{sr}\right) = (n-1)^{-2} \text{tr}\left(\mathbf{R K}_{s} \mathbf{R K}_{sr}\right),
\end{equation}
where $\mathbf{K}_s$ and $\mathbf{K}_{sr}$ are the Gram matrices, with entries $k_{r, i j} = k_{r}\left(z_{r}^{i}, z_{r}^{j}\right)$ and $k_{sr, i j} = k_{sr}\left(z_{sr}^{i}, z_{sr}^{j}\right)$. Matrix $\mathbf{R} = \mathbf{I} - \frac{1}{n} e e^{T}$, where $\mathbf{I}$ is the identity matrix and $e$ is an all-ones column vector. An inner product kernel function computes $K_r K_{sr}$.

The same HSIC measure enhances the disparity between embeddings $\mathbf{Z}_f$ and $\mathbf{Z}_{sf}$ from same adjacency matrix $\mathbf{A}_f$:
\begin{equation}
HSIC\left(\mathbf{Z}_{f}, \mathbf{Z}_{sf}\right) = (n-1)^{-2} \text{tr}\left(\mathbf{R K}_{s} \mathbf{R K}_{sf}\right),
\end{equation}

Thus, the distinct convolution loss $L_d$ is:
\begin{equation}
{L}_d = L_{dr} + L_{df} = HSIC\left(\mathbf{Z}_{r}, \mathbf{Z}_{sr}\right) + HSIC\left(\mathbf{Z}_{f}, \mathbf{Z}_{sf}\right).
\end{equation}

\textbf{Overall Objective Function.} The overall objective function, combining EL and convolution losses, is given by:
\begin{equation}
L=L_{e}+a L_{s}+b L_{d}
\end{equation}
where $a$ and $b$ are weights for the shared and distinct convolution losses, respectively.

\section{Evaluation}

This section evaluates the proposed model and presents its performance on three datasets. The implementation of our approach is based on the original codebase BLINK\footnote{https://github.com/facebookresearch/BLINK} \cite{wu2019zero} and AM-GCN\footnote{https://github.com/zhumeiqiBUPT/AM-GCN} \cite{wang2020gcn}. We compare our approach to the BLINK and SpEL\footnote{https://github.com/shavarani/SpEL} \cite{shavarani-sarkar-2023-spel} model. We selected BLINK and SpEL as baselines because of their relevance and performance benchmarks in the field. BLINK has excellent scalability and serves as part of our model's codebase. SpEL, the latest state-of-the-art as of 2023, provides a current standard for evaluating our model's improvements. Experimental details can be found in \cite{zhang_2024_12790573}

\subsection{Datasets}

Our proposed model is evaluated on three datasets which are summarized in Table~\ref{table1}. 

\begin{table}[htbp]
\centering
\caption{Summary Statistics of Datasets.}
\label{table1}
\begin{tabular}{@{}ccccc@{}}
\toprule
                                                                                       & \textbf{Train} & \textbf{Validation} & \textbf{Test} & \textbf{Entities} \\ \midrule
\textbf{\begin{tabular}[c]{@{}c@{}}Graph-TempEL:\\ Continual Entities\end{tabular}} & 1,764          & 42,096              & 48,215        & 136,227           \\
\textbf{\begin{tabular}[c]{@{}c@{}}Graph-TempEL: \\ New Entities\end{tabular}}      & 1,764          & 42,096              & 48,215        & 136,227           \\
\textbf{ZESHEL}                                                                        & 49,275         & 10,000              & 10,000        & 492,321           \\
\textbf{WikiLinksNED}                                                                  & 2,188,782      & 10,000              & 10,000        & 5,455,160         \\ \bottomrule
\end{tabular}
\end{table}

\textbf{Graph-TempEL: Continual entities and Graph-TempEL: New entities}\footnote{https://doi.org/10.5281/zenodo.12794960} are from the Graph-TempEL dataset that we constructed. Given that Wikidata5M dataset is from July 2019, to avoid temporal leakage, our dataset spans 4 years from 2019 to 2022.

Each year's data further divided into a training set (1,764), a validation set ($\approx42k$, same as original TempEL dataset), and a test set ($\approx48k$, same as original TempEL dataset). Here, the training set contains only 1,764 samples because, in the original TempEL dataset, each year's dataset contains only 1,764 `new entities' samples. The number of entities in our dataset is the same across all temporal snapshots. The data are made available at supplementary material \cite{zhang_2024_12790573}.

We also performed experiments for the full ten year period provided by TempEL while still using Wikidata5M as the reference KG. These results can be found in supplementary material \cite{zhang_2024_12790573}.

\textbf{Zero-shot Entity Linking (ZESHEL)}\footnote{https://github.com/facebookresearch/BLINK/tree/main/examples/zeshel} dataset covers various subjects, such as a fictional universe from a book or film series, mentions, and entities with detailed document descriptions. The train, validation, and test sets have 49k, 10k, and 10k samples, respectively. The entities in the validation and test sets are from domains different from those in the train set. Specifically, the training set includes domains `american football,' `doctor who,' `fallout,' `final fantasy,' `military,' `pro wrestling,' `star wars,' `world of warcraft.' The validation set includes `coronation street,' `muppets,' `ice hockey,' and `elder scrolls.'  The test set includes `forgotten realms,' `lego,' `star trek,' and `Yugioh.' This simulates the newly added entities to the knowledge graph. The number of entity candidates ranges between 10k and 100k, totaling 500k entities over all 16 domains.

\textbf{WikiLinksNED} \footnote{https://github.com/yasumasaonoe/ET4EL} dataset was created to address the challenges in the field of named entity disambiguation. Spanning a wide array of topics, from historical events to contemporary figures, the mentions and entities in this dataset are equipped with detailed document descriptions. The dataset is partitioned into train, dev, and test sets with 2.1 million, 10k, and 10k samples, respectively.

\subsection{Training Details}

We reuse the same hyperparameter settings from \cite{wu2019zero} and the same bert\_uncased\_L-8\_H-512\_A-8 pre-trained model to train the bi-encoder. The recall@$N$ is used as the evaluation metric, where $N$ equals 1, 2, 4, 8, 16, 32, and 64, respectively. If the correct answer appears within the top $N$ predictions of the model, it is considered a correct prediction. The bi-encoder is trained on the ZESHEL dataset across five epochs, utilizing 128 mentions and 128 entity tokens at a learning rate 1e-05. Conversely, the bi-encoder undergoes training for one epoch on our dataset, maintaining similar mention and entity token quantities and learning rates. The training process employs an annual training approach and tests on all test sets. More details are provided in the supplementary material \cite{zhang_2024_12790573}.

\subsection{Main Results}

Table~\ref{table2} illuminates the effectiveness of our model in mitigating temporal degradation using results derived from the Graph-TempEL dataset. Here, we use continual entities and new entities as the training set. Each column in the table represents the years' gap between the training and testing datasets, as denoted by the digits from 0 to 3. For instance, 0 implies that training and testing datasets come from the same year, while 3 indicates that the model was trained in 2019 and tested in 2022. The rows are divided based on various metrics: @1 to @64. `Boost' displays a comparison between our model TIGER and SpEL model, calculated as \( \text{Boost} = \frac{\text{Our Model's Result} - \text{Baseline Model's Result}}{\text{Baseline Model's Result}} \).

\begin{table}[t]
\centering
\caption{Temporal Degradation Mitigation Performance Across Time Gaps Using Continual Entity Samples and New Entity Samples.}
\label{table2}
\setlength{\tabcolsep}{1mm}{
\begin{tabular}{@{}cc|cccc|cccc@{}}
\toprule
                              &                     & \textbf{0}     & \textbf{1}     & \textbf{2}     & \textbf{3}     & \textbf{0}     & \textbf{1}     & \textbf{2}     & \textbf{3}     \\ \midrule
                              &                     & \multicolumn{4}{c|}{\textbf{Continual Entities}}                  & \multicolumn{4}{c}{\textbf{New Entities}}                         \\ \midrule
\multirow{4}{*}{\textbf{@1}}  & \textbf{BLINK}      & 0.177          & 0.181          & 0.182          & 0.177          & 0.132          & 0.132          & 0.132          & 0.142          \\
                              & \textbf{SpEL}       & 0.229          & 0.234          & 0.228          & 0.221          & 0.172          & 0.169          & 0.167          & 0.192          \\
                              & \textbf{TIGER}      & \textbf{0.290} & \textbf{0.292} & \textbf{0.297} & \textbf{0.304} & \textbf{0.186} & \textbf{0.195} & \textbf{0.188} & \textbf{0.217} \\
                              & \textbf{Boost (\%)} & 26.76          & 24.83          & 30.22          & 37.53          & 8.60           & 15.15          & 12.47          & 12.73          \\ \midrule
\multirow{4}{*}{\textbf{@2}}  & \textbf{BLINK}      & 0.260          & 0.265          & 0.268          & 0.263          & 0.197          & 0.197          & 0.198          & 0.211          \\
                              & \textbf{SpEL}       & 0.320          & 0.328          & 0.327          & 0.322          & 0.239          & 0.247          & 0.258          & 0.261          \\
                              & \textbf{TIGER}      & \textbf{0.404} & \textbf{0.409} & \textbf{0.414} & \textbf{0.425} & \textbf{0.274} & \textbf{0.285} & \textbf{0.277} & \textbf{0.314} \\
                              & \textbf{Boost (\%)} & 26.31          & 24.54          & 26.54          & 31.79          & 14.52          & 15.38          & 7.38           & 20.32          \\ \midrule
\multirow{4}{*}{\textbf{@4}}  & \textbf{BLINK}      & 0.357          & 0.364          & 0.367          & 0.362          & 0.277          & 0.277          & 0.278          & 0.294          \\
                              & \textbf{SpEL}       & 0.429          & 0.436          & 0.430          & 0.429          & 0.329          & 0.340          & 0.333          & 0.354          \\
                              & \textbf{TIGER}      & \textbf{0.520} & \textbf{0.524} & \textbf{0.530} & \textbf{0.543} & \textbf{0.374} & \textbf{0.389} & \textbf{0.381} & \textbf{0.421} \\
                              & \textbf{Boost (\%)} & 21.13          & 20.38          & 23.23          & 26.36          & 13.65          & 14.38          & 14.36          & 18.79          \\ \midrule
\multirow{4}{*}{\textbf{@8}}  & \textbf{BLINK}      & 0.463          & 0.469          & 0.475          & 0.470          & 0.370          & 0.370          & 0.374          & 0.392          \\
                              & \textbf{SpEL}       & 0.546          & 0.544          & 0.554          & 0.548          & 0.423          & 0.440          & 0.439          & 0.472          \\
                              & \textbf{TIGER}      & \textbf{0.628} & \textbf{0.632} & \textbf{0.637} & \textbf{0.652} & \textbf{0.483} & \textbf{0.498} & \textbf{0.490} & \textbf{0.533} \\
                              & \textbf{Boost (\%)} & 15.11          & 16.14          & 14.97          & 18.90          & 14.34          & 13.17          & 11.76          & 13.04          \\ \midrule
\multirow{4}{*}{\textbf{@16}} & \textbf{BLINK}      & 0.571          & 0.576          & 0.581          & 0.578          & 0.472          & 0.471          & 0.474          & 0.491          \\
                              & \textbf{SpEL}       & 0.645          & 0.645          & 0.656          & 0.652          & 0.539          & 0.541          & 0.554          & 0.551          \\
                              & \textbf{TIGER}      & \textbf{0.724} & \textbf{0.728} & \textbf{0.733} & \textbf{0.744} & \textbf{0.592} & \textbf{0.604} & \textbf{0.599} & \textbf{0.638} \\
                              & \textbf{Boost (\%)} & 12.36          & 12.95          & 11.76          & 14.24          & 9.74           & 11.73          & 8.22           & 15.76          \\ \midrule
\multirow{4}{*}{\textbf{@32}} & \textbf{BLINK}      & 0.675          & 0.680          & 0.685          & 0.683          & 0.576          & 0.576          & 0.577          & 0.593          \\
                              & \textbf{SpEL}       & 0.732          & 0.739          & 0.744          & 0.741          & 0.641          & 0.646          & 0.637          & 0.673          \\
                              & \textbf{TIGER}      & \textbf{0.807} & \textbf{0.809} & \textbf{0.812} & \textbf{0.821} & \textbf{0.694} & \textbf{0.704} & \textbf{0.702} & \textbf{0.732} \\
                              & \textbf{Boost (\%)} & 10.24          & 9.38           & 9.14           & 10.80          & 8.27           & 9.06           & 10.13          & 8.77           \\ \midrule
\multirow{4}{*}{\textbf{@64}} & \textbf{BLINK}      & 0.769          & 0.774          & 0.778          & 0.776          & 0.677          & 0.676          & 0.679          & 0.694          \\
                              & \textbf{SpEL}       & 0.820          & 0.827          & 0.825          & 0.824          & 0.732          & 0.733          & 0.739          & 0.754          \\
                              & \textbf{TIGER}      & \textbf{0.871} & \textbf{0.872} & \textbf{0.874} & \textbf{0.881} & \textbf{0.783} & \textbf{0.791} & \textbf{0.790} & \textbf{0.813} \\
                              & \textbf{Boost (\%)} & 6.25           & 5.43           & 5.91           & 6.86           & 7.08           & 7.85           & 6.82           & 7.84           \\ \midrule
\multicolumn{2}{c|}{\textbf{Ave. Boost (\%)}}       & 16.88          & 16.24          & 17.40          & 20.93          & 10.89          & 12.39          & 10.16          & 13.89          \\ \bottomrule
\end{tabular}}
\end{table}

\begin{figure}[t]
\centering
\includegraphics[width=0.85\linewidth]{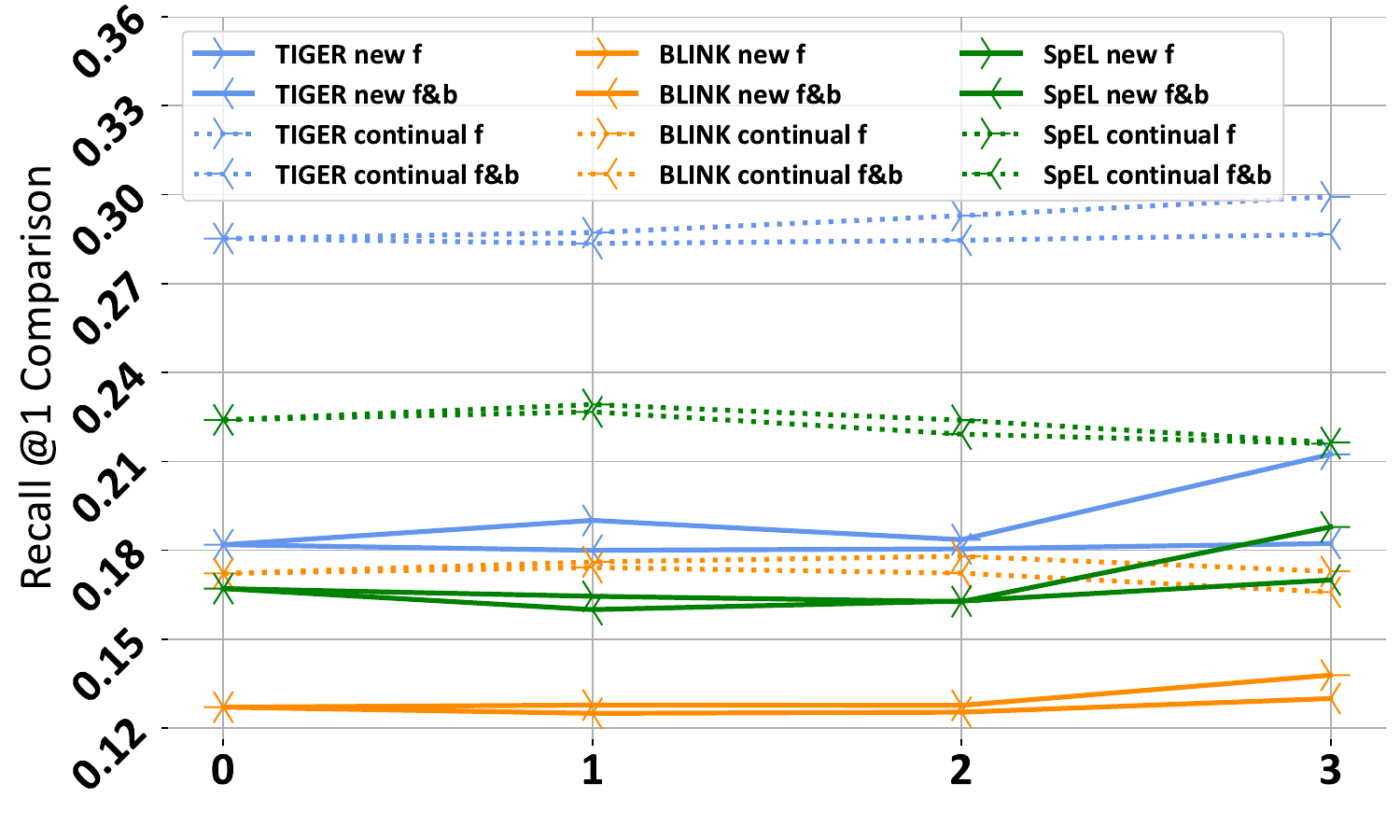}
\caption{Recall performance (recall@1) of different models on testset. The solid and dashed lines represent models training on new and continual entities.}
\label{figure4}
\end{figure}

Figure~\ref{figure4} displays recall@$N$ results from the Graph-TempEL dataset. We assessed our proposed model against the baselines using recall metrics. The x-axis indicates the year gap between training and testing sets, while the y-axis represents the recall rate. Two testing scenarios are considered: `Forward and Backward' (training on past data and testing on future data, and vice versa) and `Only Forward' (training on past data and testing on future data). For example, `Forward and Backward' averaged results from 2019 to 2022 and 2022 to 2019. The `Only Forward' scenario solely accounts for 2019 to 2022. A gap of 0 indicates identical training and testing years, making `Forward and Backward' and `Only Forward' values the same. Overall, our model consistently outperforms the baselines.

\begin{figure}[tpb]
\centering
\includegraphics[width=0.85\linewidth]{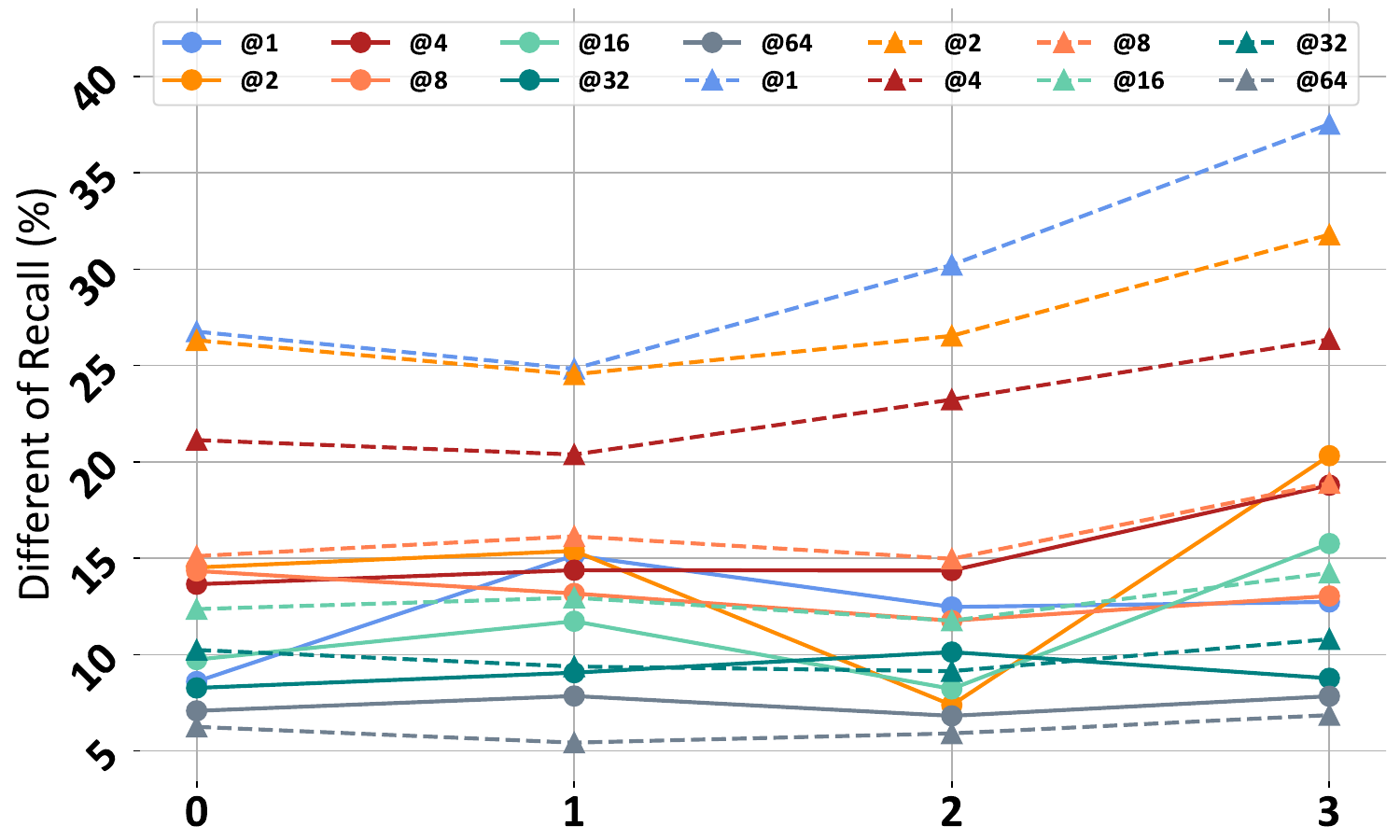}
\caption{Percentage improvement of the TIGER model compared to the SpEL model across evaluation metrics from recall@1 to recall@64.}
\label{figure5}
\end{figure}

It can be observed that compared to BLINK and SpEL, regardless of whether the training set consists of new or continual entities, our model always performs better in the `Only Forward' setting than in the `Forward and Backward' setting. This demonstrates that the model can better distinguish similar entities when graph structure is incorporated, highlighting the effectiveness of adding graph structural information. This improvement is particularly noticeable when using new entities as the training set (blue solid line), especially for larger year gap.

It is also worth noting that the improvement effect of our model diminishes gradually as the metric threshold shifts from @1 to @64, as shown in Figure~\ref{figure5}. The figure shows the `Only Forward' result. The x-axis denotes the year gap between training and testing datasets, and the y-axis represents the improvement margin of our model compared to the SpEL model. The solid line represents the model trained on `Graph-TempEL: New Entities', while the dashed line indicates the model trained on `Graph-TempEL: Continual Entities'. This figure displays results only for the only forward setting. See the supplementary material \cite{zhang_2024_12790573} for complete results of both forward and forward and backward settings. A plausible explanation for this observation is that when using the @64 threshold, the model only needs to correctly predict one out of the top 64 answers, allowing for a higher tolerance of errors. Consequently, the relative performance improvement of our model becomes less evident.

Additionally, it can be observed that the improvement of the TIGER model using continual entities as the training set (e.g., the blue dashed line in the figure) outperforms the improvement using new entities as the training set (e.g., the blue solid line). We believe this phenomenon is because continual entities, having been present in the dataset for a longer period, offer the model a richer and more consistent historical context to learn. In contrast, new entities introduce a level of uncertainty and novelty to the model, requiring it to rapidly adapt to previously unseen entities without the historical context. This may restrain the model's ability to predict accurately.

\begin{figure}[tpb]
\centering
\includegraphics[width=0.85\linewidth]{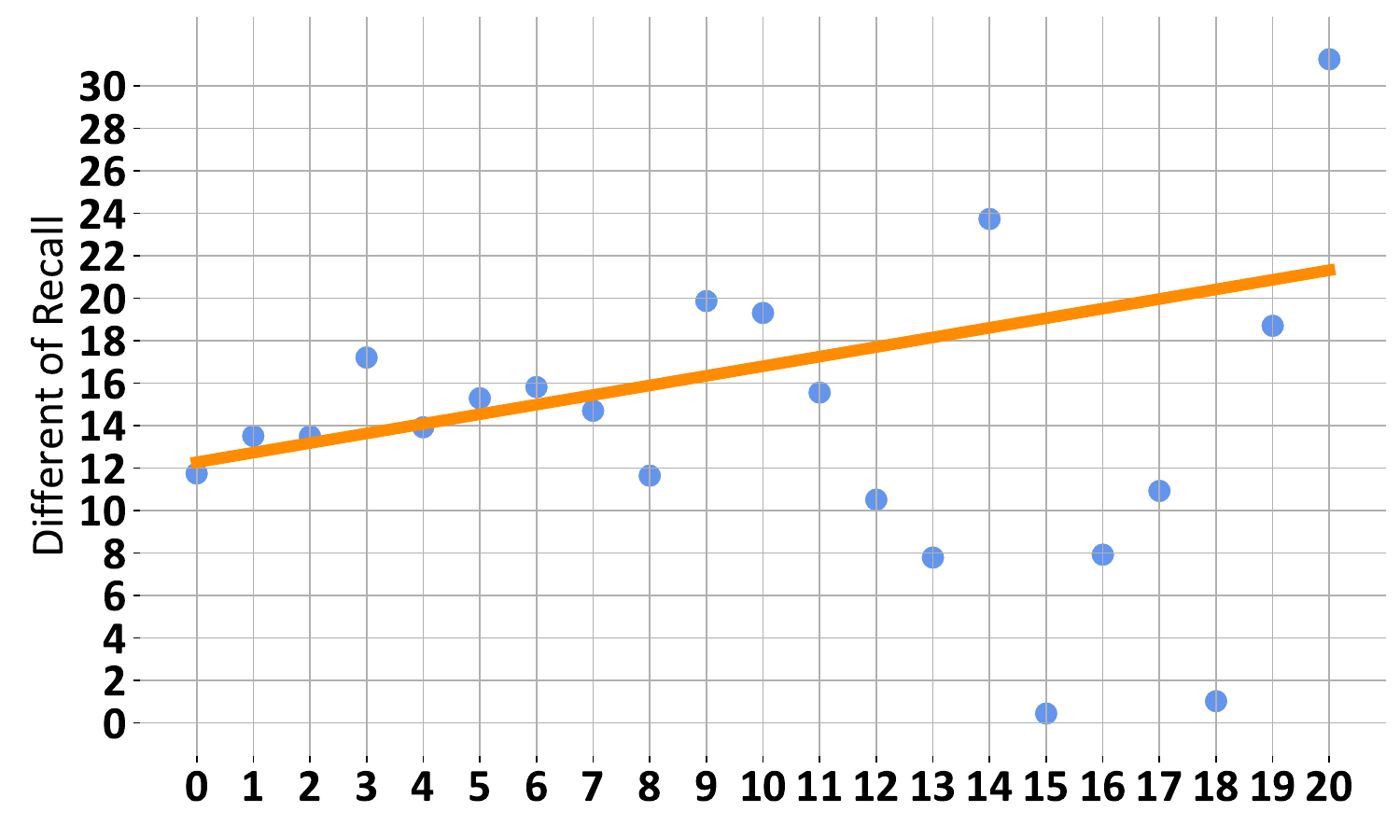}
\caption{TIGER performance improvement over the BLINK model as the degree of target entities in the relation graph increases (x-axis). The orange regression line shows a trend where TIGER achieves better performance enhancements, particularly on high-degree entities.}
\label{figure6}
\end{figure}

Figure~\ref{figure6} illustrates the improvement in recall for the TIGER model over the BLINK model in the testset across nodes of varying degrees. The x-axis represents the degree of nodes. For instance, 0 denotes isolated nodes; 1 represents nodes with only one neighbor. The y-axis indicates improvement in recall between TIGER and BLINK. Compared to the BLINK model, nodes with more neighbors provide additional information, allowing TIGER to learn more accurate node embeddings and make more precise predictions.

Table~\ref{table3} compares EL results of our model with the BLINK (biencoder and crossencoder) and SpEL (ROBERTA large) model on ZESHEL and WikilinksNED datasets. Since TIGER builds on the BLINK model by incorporating temporal and graph data optimizations, demonstrating superior performance when temporal and graph data are available (Table~\ref{table2}). Without temporal and graph data, TIGER performs similarly to the BLINK model, and SpEL remains the state-of-the-art model (Table~\ref{table3}).

\begin{table}[t]
\centering
\caption{The models' performance on non-tempral non-structure dataset.}
\label{table3}
\setlength{\tabcolsep}{0.8mm}{
\begin{tabular}{@{}cccccccc@{}}
\toprule
                                                                                             & \textbf{} & \textbf{@1} & \textbf{@4} & \textbf{@8} & \textbf{@16} & \textbf{@32} & \textbf{@64} \\ \midrule
\multirow{3}{*}{\textbf{\begin{tabular}[c]{@{}c@{}}Zeshel:\\ Forgotten Realms\end{tabular}}} & BLINK     & 0.5183      & 0.7400      & 0.7950      & 0.8375       & 0.8683       & 0.8942       \\
                                                                                             & SpEL      & \textbf{0.5717} & \textbf{0.8092} & \textbf{0.8646} & \textbf{0.8969} & \textbf{0.9373} & \textbf{0.9498}       \\
                                                                                             & TIGER     & 0.5117      & 0.7433      & 0.7983      & 0.8292       & 0.8650       & 0.8975       \\ \midrule
\multirow{3}{*}{\textbf{\begin{tabular}[c]{@{}c@{}}Zeshel: \\ Lego\end{tabular}}}            & BLINK     & 0.4170      & 0.6647      & 0.7548      & 0.8090       & 0.8599       & 0.8841       \\
                                                                                             & SpEL      & \textbf{0.4672} & \textbf{0.7216} & \textbf{0.8113} & \textbf{0.8685} & \textbf{0.9197} & \textbf{0.9420}       \\
                                                                                             & TIGER     & 0.4103      & 0.6747      & 0.7506      & 0.8098       & 0.8607       & 0.8899       \\ \midrule
\multirow{3}{*}{\textbf{\begin{tabular}[c]{@{}c@{}}Zeshel:\\ Star Trek\end{tabular}}}        & BLINK     & 0.3717      & 0.5798      & 0.6475      & 0.7052       & 0.7563       & 0.7999       \\
                                                                                             & SpEL      & \textbf{0.4316} & \textbf{0.6358} & \textbf{0.7030} & \textbf{0.7574} & \textbf{0.8122} & \textbf{0.8534}       \\
                                                                                             & TIGER     & 0.3700      & 0.5824      & 0.6485      & 0.7036       & 0.7556       & 0.7984       \\ \midrule
\multirow{3}{*}{\textbf{\begin{tabular}[c]{@{}c@{}}Zeshel:\\ Yugioh\end{tabular}}}           & BLINK     & 0.2828      & 0.4769      & 0.5504      & 0.6094       & 0.6577       & 0.6935       \\
                                                                                             & SpEL      & \textbf{0.3361} & \textbf{0.5270} & \textbf{0.6056} & \textbf{0.6615} & \textbf{0.7110} & \textbf{0.7529}       \\
                                                                                             & TIGER     & 0.2783      & 0.4798      & 0.5495      & 0.6097       & 0.6544       & 0.6935       \\ \midrule
\multirow{3}{*}{\textbf{WikilinksNED}}                                                       & BLINK     & 0.1721      & 0.4192      & 0.5467      & 0.6505       & 0.7340       & 0.7907       \\
                                                                                             & SpEL      & \textbf{0.2315} & \textbf{0.4761} & \textbf{0.5976} & \textbf{0.7084} & \textbf{0.7913} & \textbf{0.8414}       \\
                                                                                             & TIGER     & 0.1796      & 0.4227      & 0.5614      & 0.6531       & 0.7240       & 0.7973       \\ \midrule
\multirow{3}{*}{\textbf{Average}}                                                            & BLINK     & 0.3524      & 0.5761      & 0.6589      & 0.7223       & 0.7752       & 0.8125       \\
                                                                                             & SpEL      & \textbf{0.4076} & \textbf{0.6339} & \textbf{0.7164} & \textbf{0.7785} & \textbf{0.8343} & \textbf{0.8679}       \\
                                                                                             & TIGER     & 0.3500      & 0.5806      & 0.6617      & 0.7211       & 0.7719       & 0.8153       \\ \bottomrule
\end{tabular}}
\end{table}

\subsection{Qualitative Comparison}

Our model excels at accurately predicting ambiguous samples where the context is unclear or multiple interpretations exist. For example, when analyzing political events with multiple actors, our model accurately determines the correct association. In the passage
\begin{quote}
 ``\ldots Sarah Huckabee Sanders and attorney general Leslie Rutledge announced campaigns \ldots California governor Gavin Newsom was elected in 2018 with 61.9\% of the vote and is running for reelection for a second term. On September 14 2021 a \underline{recall election} was held.''   
\end{quote}
Our model correctly associates the mention ``recall election'' with the \textsc{2021 California gubernatorial recall election} entity, whereas the BLINK instead links to the \textsc{2021 Ohio 15th congressional district special election.}

Additionally, we observed that our model exhibits a higher prediction accuracy for samples related to temporal aspects. 

For example, in the passage:
\begin{quote}
    ``Dundalk entered the 2021 season as the \underline{FAI Cup} holders, and were still the League of Ireland Cup holders from 2019 \ldots''
\end{quote}
our model correctly identified the mention ``FAI Cup'' as referring to the \textsc{2021 FAI Cup} entity whereas the BLINK linked to the \textsc{2009–10 in Scottish football} entity.

\section{Conclusion and Future Work}

This paper introduces \textbf{TIGER}, a \textbf{T}emporally \textbf{I}mproved \textbf{G}raph \textbf{E}ntity \textbf{L}inke\textbf{r}, to address temporal degradation. By adaptively combining the distinct and shared features between different entity relationships, the model is able to ensure that the semantic differences between different entities remain intact and do not diminish over time. We expanded the TempEL dataset by incorporating yearly entity relationships from the Wikidata5M dataset, creating Graph-TempEL, which enhances its suitability for studying temporal degradation. The dataset provides four yearly snapshots from 2019 to 2022. Each snapshot features entity descriptions, mention contexts, structure and feature graphs, and an entity feature matrix. Experiments on Graph-TempEL dataset show that our model can effectively prevent temporal degradation, demonstrating a 16.24\% performance boost over the state-of-the-art in a temporal setting when the time gap is one year and an improvement to 20.93\% as the gap expands to three years. Going forward, we see a few areas of future work:

\textbf{Contrastive Learning.} Contrastive learning has been extensively applied to graph neural networks in recent research, yet its application to temporal datasets remains limited. When a dataset contains snapshots from different years, some entities will likely appear in multiple snapshots. If certain relationships between entities repeatedly occur in various snapshots, these entity pairs can be assumed to have stronger connections. Such pairs can then serve as high-quality positive samples. Conversely, entity pairs that had relationships that subsequently disappeared can serve as negative samples. With these high-quality positive and negative samples, the model's performance can potentially be improved under unsupervised conditions.


\textbf{Multilingual Entity Linking.} Despite language differences, relationships may share similarities, allowing for multilingual entity linking. While existing methods like \cite{10.1145/3543873.3587627} concentrate on multilingual aligned embedding and community detection in graphs, there is limited research addressing the issue of temporal degradation in multilingual contexts. Our work is currently limited to English, with low-resource datasets not covered. If multilingual entity descriptions and entity relationships exist at time $t_1$, the model could benefit from these diverse language resources, thereby further improving the accuracy at time $t_2$ and better preventing temporal degradation.



\begin{ack}
The first author is supported by the China Scholarship Council (NO. 202206540007) and the University of Amsterdam. This funding source had no influence on the study design, data collection, analysis, or manuscript preparation and approval. This work is partially supported by the EU’s Horizon Europe programme, in the ENEXA project (grant Agreement no. 101070305).
\end{ack}



\bibliography{m511}

\end{document}